\documentclass[manuscript,sigconf,natbib=false]{acmart}
\AtBeginDocument{%
  }

\setcopyright{acmlicensed}
\copyrightyear{2026}
\acmYear{2026}
\acmDOI{XXXXXXX.XXXXXXX}
\acmConference[BioKDD '26]{the 25th International Workshop on Data Mining in Bioinformatics}{August 09--13,
  2026}{Jeju, Korea}

\RequirePackage[
  datamodel=acmdatamodel,
  style=acmnumeric,
  ]{biblatex}

\usepackage{booktabs}
\usepackage{multirow}
\usepackage[table]{xcolor}
\definecolor{revisionblue}{HTML}{0057B8}

\newcommand{\val}[2]{#1{\scriptsize$\pm$#2}}
\newcommand{\topval}[2]{\cellcolor{red!12}\textbf{#1}{\scriptsize$\pm$#2}}
\newcommand{\secondval}[2]{\cellcolor{blue!8}\textbf{#1}{\scriptsize$\pm$#2}}
\newcommand{\bestrank}[1]{\textbf{#1}}

\begin{document}

\title[Geometry-Informed PEFT for BBBP Prediction]{{Geometry}-Informed Parameter-Efficient Fine-Tuning of Pre-trained Molecular GNNs for Blood--Brain Barrier Permeability Prediction}


\author{Marco Vieto Vega} \authornote{Both authors contributed equally to this research.} 
\email{vietomarc@myvuw.ac.nz}
\affiliation{%
  \institution{School of Mathematics and Statistics \\ Victoria University of Wellington}
  \city{Wellington}
  \country{New Zealand}
}

\author{Long D. Nguyen} \authornotemark[1] \email{duylong.nguyen@vuw.ac.nz} 
\affiliation{%
  \institution{School of Mathematics and Statistics \\ Victoria University of Wellington}
  \city{Wellington}
  \country{New Zealand}
}

\author{Binh P. Nguyen} \authornote{Corresponding author.} \email{binh.p.nguyen@vuw.ac.nz}
\affiliation{%
  \institution{School of Mathematics and Statistics \\ Victoria University of Wellington}
  \city{Wellington}
  \country{New Zealand}
}

\renewcommand{\shortauthors}{Marco et al.}


\begin{abstract}
Blood--brain barrier permeability (BBBP) prediction is a critical screening task in central nervous system drug discovery, where candidate molecules must be assessed according to whether they can cross, or should be prevented from crossing the blood--brain barrier. However, this task remains challenging due to limited, class-imbalanced datasets and sensitivity to molecular structure. Recent advances in deep learning have established graph neural networks (GNNs) as a powerful approach for molecular representation learning, and pre-trained molecular GNNs provide transferable knowledge for downstream tasks. However, full fine-tuning of pre-trained models is often parameter-inefficient and prone to overfitting, while existing parameter-efficient fine-tuning (PEFT) methods mainly adapt node features or the two-dimensional covalent graph, limiting their ability to capture three-dimensional geometry and second-order interactions.
To address these limitations, we propose \textbf{BBBP-GeoPEFT}, a geometry-informed PEFT framework for pre-trained molecular GNNs. BBBP-GeoPEFT constructs distance-based graphs at multiple cutoffs and
their corresponding line graphs from molecular conformers to capture
spatial atom interactions and second-order edge interactions.
Lightweight auxiliary geometric graph encoders produce cutoff-specific
representations, which are incorporated into each pre-trained layer
through node-wise cutoff attention and gated residual connections. This design preserves pre-trained knowledge while incorporating permeability-relevant geometric information with a small trainable-parameter budget.
Experiments on a curated BBBP dataset demonstrate that BBBP-GeoPEFT achieves competitive and consistent performance compared with full fine-tuning and representative PEFT baselines. Under both random and scaffold splitting settings, BBBP-GeoPEFT achieves competitive or improved ROC-AUC and Accuracy in most experiments while updating only 10.1\% of the model parameters.
\end{abstract}



\maketitle

\section{Introduction}

Molecular property prediction plays a central role in drug discovery and materials science~\cite{yang2019analyzing}. Among these tasks, blood--brain barrier permeability (BBBP) prediction is particularly important for central nervous system drug development, where candidate molecules must be evaluated based on their ability to cross the blood--brain barrier. The blood--brain barrier is a highly selective biological interface that regulates the transport of compounds into the brain, protecting the central nervous system and maintaining homeostasis~\cite{Bradbury1993,Segarra2021}. Compounds targeting neurological diseases require sufficient permeability to reach their targets, while for many non-central nervous system drugs, low permeability is desirable to avoid unintended side effects. Experimental measurement of BBBP through in vitro~\cite{Pardridge2003} or in vivo assays~\cite{Palmer2013} is costly and time-consuming, making computational prediction an essential tool for early-stage screening. However, BBBP prediction remains challenging due to limited and class-imbalanced datasets, as well as the strong dependence of permeability on complex molecular structure.

Traditional approaches for BBBP prediction rely on physicochemical rules or classical machine learning models built on handcrafted molecular descriptors~\cite{Wager2010,Ghose2016,Gupta2019}. While these methods are interpretable and effective in low-data settings, they are limited in their ability to capture complex structural patterns. Recently, graph neural networks (GNNs) have emerged as a powerful framework for molecular representation learning by modeling molecules as graphs of atoms and bonds~\cite{zhu2022hignn,nguyen2024smiles}. Furthermore, self-supervised pre-training enables GNNs to learn transferable representations from large unlabeled molecular datasets, improving performance on downstream molecular property prediction tasks~\cite{Hu*2020Strategies,you2020graph,xia2022simgrace}.

Despite these advances, adapting pre-trained molecular GNNs to BBBP prediction remains challenging. Fully fine-tuning pre-trained models requires updating a large number of parameters, which can lead to overfitting and catastrophic forgetting, especially when labeled data are scarce or imbalanced~\cite{zhou2021overcoming}. Parameter-efficient fine-tuning (PEFT) methods address this issue by introducing lightweight task-specific modules while freezing most pre-trained parameters~\cite{Li2024,papageorgiou2025graph,fang2023universal,huang2025one,fu2025edge}. However, existing PEFT approaches for molecular graphs are limited in their ability to exploit structural information relevant to BBBP. Prompt-based methods mainly modify input representations and often underperform compared to full fine-tuning~\cite{Li2024}. Feature-based adapters operate primarily on node features, while recent structure-aware adapters remain restricted to the original 2D covalent graph. Consequently, these methods do not explicitly incorporate
conformer-derived geometry or second-order relations between spatial
atom-pair interactions that may be relevant to molecular permeability.

To address this limitation, we propose \textbf{BBBP-GeoPEFT}, a geometry-informed parameter-efficient fine-tuning framework for BBBP prediction. BBBP-GeoPEFT augments a pre-trained molecular GNN with geometric
representations derived from RDKit-generated conformers. Specifically,
we construct distance-based graphs under multiple spatial cutoffs to
model atom proximity and derive corresponding line graphs to represent
second-order edge interactions. Lightweight auxiliary geometric graph
encoders produce a cutoff-specific representation from each graph pair.
At every backbone layer, node-wise cutoff attention aggregates these
representations and forms a geometric adaptation residual that is
combined with the pre-trained state through gated connections. This
design incorporates conformer-derived information while preserving
pre-trained knowledge and maintaining parameter efficiency.
We evaluate BBBP-GeoPEFT on a curated BBBP dataset derived from the Blood--Brain Barrier Database. Experimental results show that BBBP-GeoPEFT achieves strong performance across multiple evaluation metrics, outperforming or remaining competitive with full fine-tuning and representative parameter-efficient baselines while using only a small fraction of trainable parameters. These results demonstrate that incorporating multi-scale conformer geometry into parameter-efficient adaptation is a promising direction for BBBP prediction.

\noindent\textbf{Contributions.} 
Our main contributions are summarized as follows:
\begin{itemize}
    \item We propose BBBP-GeoPEFT, a geometry-informed parameter-efficient
    fine-tuning framework that augments a largely frozen molecular GNN
    with representations derived from multi-scale conformer geometry.
    
    \item We introduce a multi-scale graph construction based on distance graphs
    and their corresponding line graphs, enabling the model to represent
    spatial atom interactions and second-order edge interactions across
    multiple cutoffs.
    
    \item We design a node-wise cutoff attention module that aggregates
    cutoff-specific geometric representations and augments each backbone
    layer through a gated geometric adaptation residual.
    
    \item We conduct extensive experiments on a curated BBBP dataset, demonstrating that BBBP-GeoPEFT achieves competitive or improved performance while maintaining parameter efficiency compared to full fine-tuning and existing parameter-efficient methods.
\end{itemize}

\newpage

\section{Related Work}

\subsection{Computational Blood--Brain Barrier Permeability Prediction}

Computational approaches for blood--brain barrier permeability prediction have evolved from rule-based and statistical models to advanced machine learning and deep learning methods. Early studies focused on physicochemical descriptors and statistical modeling techniques. Linear Free Energy Relationship models~\cite{Platts2001} based on multivariate linear regression were used to correlate permeability with fundamental molecular properties such as hydrogen bonding, polarizability, and molecular size. Similarly, Linear Discriminant Analysis~\cite{Vilar2010} has been applied to construct binary classifiers using simple two-dimensional descriptors, including lipophilicity, topological polar surface area, and counts of acidic and basic atoms.

To overcome the limitations of fixed thresholds, multiparameter optimization frameworks were introduced to integrate multiple physicochemical properties into unified scoring functions. Representative methods include the CNS MPO score~\cite{Wager2010}, which combines several molecular descriptors through desirability functions, and its extensions such as TEMPO~\cite{Ghose2016} and BBB Score~\cite{Gupta2019}, which employ optimization techniques and piecewise functions to assign probabilistic permeability scores. While these approaches are interpretable and computationally efficient, they rely heavily on predefined descriptors and may fail to capture complex structural dependencies.

With the increasing availability of data and computational resources, conventional machine learning methods such as k-Nearest Neighbors (k-NN)~\cite{Cover1967}, Support Vector Machines (SVM)~\cite{Cortes1995}, Random Forests (RF)~\cite{Breiman2001}, and XGBoost~\cite{Chen2016} have been widely adopted for BBBP prediction. These models are typically trained on handcrafted features such as quantitative structure--activity relationship descriptors or molecular fingerprints. More recently, deep learning methods have demonstrated improved performance by learning representations directly from molecular structure. Ensemble architectures, such as iBBBP-Ensemble~\cite{NguyenVo2024}, combine Convolutional Neural Networks and Multilayer Perceptrons to integrate multiple feature modalities. In addition, pre-trained graph-based models, such as Masked Graph Transformer Encoders~\cite{Vinh2025}, further enhance representation learning by leveraging large-scale unlabeled molecular data.

Despite these advances, existing BBBP prediction methods exhibit two key limitations. First, many approaches rely on fixed feature representations, which restrict their ability to capture complex molecular structure. Second, deep learning models often require training or full fine-tuning, which is challenging under limited and imbalanced datasets. 
Moreover, most methods focus on 2D molecular representations and do not
explicitly incorporate conformer-derived spatial relations or
second-order dependencies among atom-pair interactions, which may provide
complementary information for molecular permeability prediction.
These limitations motivate the need for more flexible and structure-aware adaptation strategies.

\subsection{Efficient Adaptation of Pre-trained Graph Models}

Beyond task-specific modeling, another line of research focuses on improving the adaptation of pre-trained graph neural networks through parameter-efficient fine-tuning. Instead of updating all parameters, PEFT methods introduce a small number of trainable components while keeping the backbone largely frozen, reducing computational cost and mitigating overfitting.

Existing approaches can be broadly categorized into \emph{graph prompt tuning} and \emph{adapter-based tuning}. Prompt-based methods adapt GNNs by modifying the input graph or task formulation without changing the model architecture. Early approaches, such as GPF and GPF-plus~\cite{fang2023universal}, introduce learnable feature-level prompts appended to node representations. Subsequent works extend prompting to structural information: EdgePrompt~\cite{fu2025edge} injects prompts into edges, while GraphTOP~\cite{fu2025graphtop} and UniPrompt~\cite{huang2025one} modify graph topology to generate task-specific structural signals. More unified frameworks, such as GPPT~\cite{sun2022gppt} and All-in-One~\cite{sun2023all}, reformulate downstream tasks into a common prompting paradigm.

Adapter-based methods introduce lightweight trainable modules within pre-trained GNNs. AdapterGNN~\cite{Li2024} proposes graph-specific adapter designs with parallel modules and learnable scaling, while G-Adapter~\cite{Gui2024} and GConv-Adapter~\cite{papageorgiou2025graph} incorporate graph convolution into adapters to capture structural information. In addition, low-rank adaptation techniques such as LoRA~\cite{hu2022lora} reduce parameter overhead by factorizing weight updates.

While these methods improve adaptation efficiency, most derive their
trainable corrections from node features or the original local graph
connectivity. They therefore rarely exploit conformer-derived spatial
relations or second-order edge interactions during adaptation. This
limitation motivates a parameter-efficient mechanism that supplies
geometric graph representations to the pre-trained backbone.

\subsection{Geometry-Aware and Higher-Order Molecular Learning}

Orthogonal to adaptation efficiency, recent work has explored incorporating geometric and higher-order structural information into molecular representation learning beyond traditional 2D graphs. Early approaches, such as Mol-GDL~\cite{shen2023molecular}, construct multiple graphs under different distance thresholds, enabling models to capture both covalent and non-covalent interactions from three-dimensional geometry. 
A line of research further develops simplicial and spectral methods based on Hodge theory. For example, HL-HGAT~\cite{huang2025hl} applies Hodge-Laplacian filtering with efficient polynomial approximation, while HLSAD~\cite{Frantzen2025} leverages spectral properties of Hodge Laplacians to capture structural evolution in higher-order networks. Other works, such as TorGNN~\cite{Shen2025}, incorporate advanced topological invariants into message passing to model higher-order interactions. AGBT~\cite{chen2021algebraic} combines algebraic graph fingerprints with transformer representations learned from SMILES strings, while topological fusion networks extract simplicial features from three-dimensional structures and integrate them into attention mechanisms.

In addition, multi-modal and optimization-based approaches~\cite{dong2025exploring} integrate 2D, 3D, and topological features, while explicit topological descriptors provide efficient alternatives based on handcrafted invariants. More recent works, such as ENINet~\cite{Mao2025}, utilize line graphs to explicitly model equivariant many-body interactions, and MatRIS~\cite{zhou2026matris} captures three-body interactions through line graph representations, further improving expressiveness in modeling complex molecular systems.

Despite their effectiveness, these methods typically require training specialized architectures or directly integrating geometric and topological features into the backbone model. As a result, they are not designed for parameter-efficient adaptation of pre-trained graph neural networks, and their computational cost can limit scalability in practical settings. This highlights the need for approaches that can leverage geometric and higher-order information while maintaining efficiency.

\paragraph{\textbf{Discussion}.}
Existing research on BBBP prediction, parameter-efficient fine-tuning, and geometry-aware molecular learning has made substantial progress along three complementary directions. However, these lines of work remain largely disconnected. BBBP-specific models either rely on handcrafted descriptors or require full model training, without leveraging efficient adaptation of pre-trained representations. Parameter-efficient fine-tuning methods focus on reducing computational cost but are typically limited to 2D graph representations. Meanwhile, geometry-aware and higher-order models capture rich structural interactions but often require specialized architectures and are not designed for adapting pre-trained models.

{Our contribution is a geometry-aware PEFT design that combines established structural and adaptation components. 
It augments a largely frozen 2D molecular GNN with cutoff-specific
representations obtained from conformer-derived distance and line graphs.
}

\section{Preliminaries}

\paragraph{Molecular Graph and Geometry.}
We represent a molecule as an attributed graph $\mathcal{G}=(\mathcal{V},\mathcal{E},{\mathbf{A}},\mathbf{X})$, where $\mathcal{V}=\{v_1,\ldots,v_n\}$ denotes atoms, $\mathcal{E}\subseteq \mathcal{V}\times\mathcal{V}$ denotes chemical bonds, and $\mathbf{X}\in\mathbb{R}^{n\times d_x}$ is the node feature matrix. The graph structure is encoded by an adjacency matrix $\mathbf{A}\in\{0,1\}^{n\times n}$, where each node $v_i$ is associated with a neighborhood $\mathcal{N}(v_i)$. Edges may further carry attributes $\mathbf{e}_{ij}\in\mathbb{R}^{d_e}$, collectively denoted as $\mathbf{E}$.

In addition to the 2D molecular graph, each molecule is associated with 3D atomic coordinates $\mathbf{R} \in \mathbb{R}^{|\mathcal{V}| \times 3}$ obtained from conformers. These coordinates provide geometric information, such as inter-atomic distances and spatial arrangements, which are not captured by the covalent graph.

\paragraph{BBBP Prediction Task.}
Given a molecule represented by $(\mathbf{A}, \mathbf{X}, \mathbf{E})$ and its corresponding label $y \in \{0,1\}$, the goal of blood--brain barrier permeability prediction is to determine whether the molecule can penetrate the blood--brain barrier. We denote $y=1$ as BBB-permeable and $y=0$ as non-permeable. The model outputs a prediction $\hat{y}\in[0,1]$, which is trained using a supervised loss function $\mathcal{L}(\hat{y}, y)$ over a dataset $\mathcal{D}$.

\paragraph{Pre-trained GNN and Downstream Molecular Task.}
Let $f_{\boldsymbol{\theta}}$ denote a pre-trained graph neural network encoder operating on the molecular graph, and let $g_{\boldsymbol{\phi}}$ denote a task-specific prediction head. Given an input molecule, the model produces a prediction:
\begin{equation}
\hat{y} = g_{\boldsymbol{\phi}}\!\left(f_{\boldsymbol{\theta}}(\mathbf{A},\mathbf{X},\mathbf{E})\right),
\end{equation}
where $\hat{y}$ is the predicted probability. Pre-training enables $f_{\boldsymbol{\theta}}$ to capture general molecular representations from large unlabeled datasets.

\paragraph{Parameter-Efficient Adaptation.}
To adapt pre-trained models under limited labeled data, we adopt a parameter-efficient fine-tuning setting. Instead of updating all parameters, we partition the encoder into frozen parameters $\boldsymbol{\theta}_f$ and a small set of tunable parameters $\boldsymbol{\theta}_t$, and introduce additional lightweight adaptation parameters $\boldsymbol{\psi}$. The adapted model $\tilde{f}$ is defined as:
\begin{equation}
\hat{y} = g_{\boldsymbol{\phi}}\!\left(\tilde{f}_{\boldsymbol{\theta}_f, \boldsymbol{\theta}_t, \boldsymbol{\psi}}(\mathbf{A},\mathbf{X},\mathbf{E}, \mathbf{R})\right),
\end{equation}
and is optimized by:
\begin{equation}
\min_{\boldsymbol{\theta}_t, \boldsymbol{\psi}, \boldsymbol{\phi}}
\frac{1}{|\mathcal{D}|}
\sum_{(\mathbf{A},\mathbf{X},\mathbf{E},\mathbf{R},y)\in\mathcal{D}}
\mathcal{L}\!\left(
g_{\boldsymbol{\phi}}\!\left(\tilde{f}_{\boldsymbol{\theta}_f, \boldsymbol{\theta}_t, \boldsymbol{\psi}}(\mathbf{A},\mathbf{X},\mathbf{E}, \mathbf{R})\right), y
\right),
\end{equation}
where $|\boldsymbol{\theta}_t| + |\boldsymbol{\psi}| \ll |\boldsymbol{\theta}|$. This formulation enables efficient adaptation while preserving pre-trained knowledge.

\paragraph{Conformer-Derived Graph Construction.}
To incorporate structural information beyond the 2D graph, we utilize the 3D coordinates $\mathbf{R}$ to construct auxiliary graph
representations. 
Distance-based
graphs encode spatial proximity between atoms, while their corresponding
line graphs represent relations between atom-pair interactions that share
an atom. Constructing these graph pairs at multiple cutoffs provides
complementary conformer-derived representations for adapting the
pre-trained GNN.

\section{Method}

\subsection{Overview}

\begin{figure*}[!ht]
    \centering
    \Description{BBBP-GeoPEFT overview figure.}
    \includegraphics[width=\textwidth]{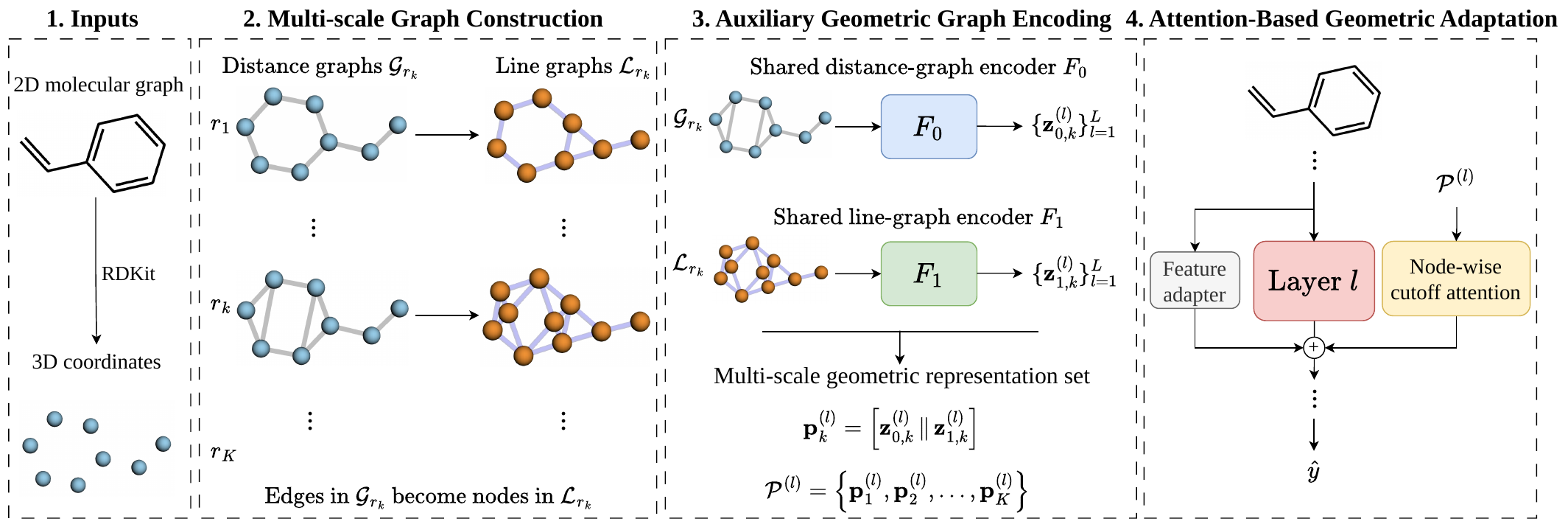}
    \caption{
    Overview of BBBP-GeoPEFT. 
    Given a molecule, the original 2D molecular graph is processed by a pre-trained GNN, while 3D coordinates $\mathbf{R}$ are used to construct multi-scale graph views. Specifically, distance-based graphs $\{\mathcal{G}_{r_k}\}_{k=1}^{K}$ are built under multiple cutoffs, and their corresponding line graphs $\{\mathcal{L}_{r_k}\}_{k=1}^{K}$ represent second-order edge interactions.
    A shared distance-graph encoder $F_0$ and line-graph encoder $F_1$ process the two graph families. 
    At layer \(l\), their pooled outputs are
    concatenated into cutoff-specific geometric representations
    \(\mathbf{p}^{(l)}_k\), yielding the multi-scale geometric representation
    set \(\mathcal{P}^{(l)}\). Node-wise cutoff attention assigns
    node-dependent weights to these representations and produces a geometric
    adaptation residual. A feature adapter supplies an additional
    feature-level residual, and gated connections combine both residuals
    with the pre-trained node state. Graph-level readout and an MLP produce
    the final prediction \(\hat{y}\).
    }
\label{fig:overview}
\end{figure*}

We propose \textbf{BBBP-GeoPEFT} (Figure~\ref{fig:overview}), a parameter-efficient framework that augments a pre-trained molecular graph neural network with multi-scale graph representation derived from 3D geometry.
Given a molecule, the pre-trained GNN processes the original 2D
molecular graph, while the 3D coordinates are used to construct
distance-based graphs at multiple cutoffs and their corresponding line
graphs. The distance graphs model spatial atom proximity, whereas the
line graphs represent second-order edge interactions.
Shared auxiliary geometric graph encoders transform these
graph pairs into cutoff-specific geometric representations. At each
backbone layer, node-wise cutoff attention aggregates the representations
according to the current node state and produces a geometric adaptation
residual. The residual is incorporated through a gated update, allowing
the model to use conformer-derived information while training only a
small subset of parameters.

\subsection{Multi-scale Graph Construction}


We construct distance and line graphs from three-dimensional atomic
coordinates to capture interactions beyond the covalent graph.

\paragraph{Distance-based graphs.}
Given atomic coordinates $\mathbf{R}\in\mathbb{R}^{n\times 3}$, we define a distance-based graph under cutoff $r$ as $\mathcal{G}_r=(\mathcal{V},\mathcal{E}_r,\mathbf{X}_r)$, where:
\begin{equation}
\mathcal{E}_r = \{(i,j) \mid i \neq j, \|\mathbf{R}_i - \mathbf{R}_j\|_2 \le r\}.
\end{equation}
The node features $\mathbf{X}_{r}$ are initialized using the same atom-level features as the original 2D molecular graph. This graph captures spatial proximity and includes both bonded and non-covalent interactions.


\paragraph{Line graphs for second-order edge interactions.}
To model relations among spatial atom-pair interactions, we construct the
line graph
\(\mathcal{L}_{r}
=
(\mathcal{V}_{r}^{\ell},
\mathcal{E}_{r}^{\ell},
\mathbf{X}_{r}^{\ell})\),
where each line-graph node corresponds to an edge of
\(\mathcal{G}_{r}\):
\begin{equation}
\mathcal{V}_{r}^{\ell}=\mathcal{E}_{r}.
\end{equation}
Two distinct nodes in \(\mathcal{L}_{r}\) are connected when their
corresponding edges in \(\mathcal{G}_{r}\) share an atom:
\begin{equation}
\mathcal{E}_{r}^{\ell}
=
\left\{
(e_{ij},e_{uv})
\;\middle|\;
e_{ij}\neq e_{uv},
\;
\{i,j\}\cap\{u,v\}\neq\emptyset
\right\}.
\end{equation}
We refer to each such relation between two incident spatial edges as a
\emph{second-order edge interaction}. Each line-graph node associated
with an edge \(e_{ij}\in\mathcal{E}_{r}\) is initialized by applying a
radial basis function encoding to the corresponding inter-atomic
distance:
\begin{equation}
\mathbf{x}_{ij}^{\ell}
=
\textsc{RBF}
\left(
\left\|
\mathbf{R}_{i}-\mathbf{R}_{j}
\right\|_{2}
\right).
\end{equation}
Message passing over \(\mathcal{L}_{r}\) therefore models dependencies
between spatial edges that share a common atom.

\paragraph{Multi-scale construction.}
We use multiple cutoff values $\mathcal{R}=\{r_1,\dots,r_K\}$ to construct:
\begin{equation}
\{\mathcal{G}_{r_k}\}_{k=1}^K, \quad \{\mathcal{L}_{r_k}\}_{k=1}^K.
\end{equation}
Different cutoffs define geometric neighborhoods over varying spatial
ranges. Smaller cutoffs emphasize local atom proximity, whereas larger
cutoffs include longer-range spatial relations. The family
\(\{\mathcal G_{r_k}\}_{k=1}^{K}\) represents cutoff-dependent atom connectivity,
whereas \(\{\mathcal L_{r_k}\}_{k=1}^{K}\) represents second-order edge
interactions induced by shared atoms. Together, the two graph families
provide multi-scale conformer geometry for parameter-efficient
adaptation.

\subsection{Auxiliary Geometric Graph Encoding}

We encode the multi-scale distance and line graphs using lightweight
auxiliary geometric graph encoders.

\paragraph{Shared graph encoders.}
\label{para:structural}
Instead of assigning independent encoders to each cutoff, we use two shared GNN encoders: \(F_0(\cdot)\) for distance-based graphs and \(F_1(\cdot)\) for line graphs. 
For each cutoff \(r_k\), the layer-wise updates are:
\begin{align}
\mathbf{S}_{0,k}^{(l)}
&=
F_0^{(l)}(\mathbf{S}_{0,k}^{(l-1)}),\\
\mathbf{S}_{1,k}^{(l)}
&=
F_1^{(l)}(\mathbf{S}_{1,k}^{(l-1)}).
\end{align} 
Equivalently, for all scales, the shared encoders produce:
\begin{align}
    \{\{\mathbf{S}_{0,k}^{(l)}\}_{k=1}^{K}\}_{l=1}^{L}
&=
F_0^{(l)}\left(\{\mathcal{G}_{r_k}\}_{k=1}^{K}\right), \\
\{\{\mathbf{S}_{1,k}^{(l)}\}_{k=1}^{K}\}_{l=1}^{L}
&=
F_1^{(l)}\left(\{\mathcal{L}_{r_k}\}_{k=1}^{K}\right).
\end{align}
The encoders $F_0$ and $F_1$ are shared across all cutoff values, so the model learns scale-consistent structural transformations while reducing the parameter cost from scale-specific encoders to only two encoder families. 
The distance-graph encoder models cutoff-dependent atom
connectivity, whereas the line-graph encoder models second-order
relations between spatial edges that share an atom.

\paragraph{Parameter-efficient design.}
To reduce the number of trainable parameters, we employ a low-rank factorized parameterization for the MLP layers within the shared auxiliary graph encoders. Specifically, while the message-passing mechanism of the GNN remains unchanged, we apply low-rank decomposition only to the linear transformations inside the MLP blocks.
Instead of learning full weight matrices in $\text{MLP}^{(l)}$, each linear transformation is represented using a low-rank factorization. Concretely, for a linear layer with weight $\mathbf{W}^{(l)} \in \mathbb{R}^{d \times d}$, we parameterize it as:
\begin{equation}
\mathbf{W}^{(l)} = \mathbf{B}^{(l)} \mathbf{A}^{(l)},
\end{equation}
where $\mathbf{A}^{(l)} \in \mathbb{R}^{r \times d}$ and $\mathbf{B}^{(l)} \in \mathbb{R}^{d \times r}$ are trainable low-rank matrices with $r \ll d$. This factorized parameterization reduces the number of trainable parameters from $\mathcal{O}(d^2)$ to $\mathcal{O}(2rd)$, i.e., $\mathcal{O}(rd)$.
Importantly, this low-rank design is applied only to the MLP components of
the auxiliary graph encoders, while the neighborhood aggregation operations
remain unchanged. Most parameters of the pre-trained backbone are frozen,
with only its batch-normalization parameters fine-tuned during downstream
adaptation.
This preserves the capacity of the auxiliary encoders to model the
distance and line graphs while improving parameter efficiency.

\paragraph{Multi-scale geometric representation set.}
We aggregate distance-graph and line-graph representations using a permutation-invariant pooling function:
\begin{equation}
\mathbf{z}_{0,k}^{(l)} = \textsc{Pool}(\mathbf{S}_{0,k}^{(l)}), \quad
\mathbf{z}_{1,k}^{(l)} = \textsc{Pool}(\mathbf{S}_{1,k}^{(l)}).
\end{equation}

For each cutoff \(r_k\), we obtain a cutoff-specific geometric
representation by concatenating the pooled distance-graph and line-graph
representations:
\begin{equation}
\mathbf{p}_k^{(l)} = [\mathbf{z}_{0,k}^{(l)} \,\|\, \mathbf{z}_{1,k}^{(l)}].
\end{equation}
The set $\mathcal{P}^{(l)} = \{\mathbf{p}_k^{(l)}\}_{k=1}^K$ forms a multi-scale geometric representation set that summarizes the
conformer across the selected distance cutoffs. In practice, the hidden dimension of each auxiliary graph encoder is set
to half the hidden dimension of the pre-trained encoder.

\subsection{Attention-Based Geometric Adaptation}

We integrate multi-scale geometric representation set into the pre-trained GNN through a node-wise cutoff attention module applied at each layer.

\paragraph{Node-wise cutoff attention.}
Let \(\mathbf{h}^{(l)}_i\) denote the representation of node \(i\) at
layer \(l\). We compute its similarity to every cutoff-specific
geometric representation in \(\mathcal{P}^{(l)}\). The node state acts
as the query, while each element of \(\mathcal{P}^{(l)}\) corresponds
to one distance cutoff:
\begin{equation}
s_{i,j}^{(l)} =
\frac{\langle \mathbf{h}_i^{(l)}, \mathbf{p}_j^{(l)} \rangle}{\sqrt{d}}.
\end{equation}
The attention weights are computed as:
\begin{equation}
\alpha_{i,j}^{(l)} = \textsc{Softmax}_j(s_{i,j}^{(l)}),
\end{equation}
and the geometric adaptation residual is defined as:
\begin{equation}
\Delta \mathbf{h}_{i,\mathrm{geo}}^{(l)} =
\sum_{j=1}^K \alpha_{i,j}^{(l)} \mathbf{p}_j^{(l)}.
\end{equation}
This mechanism enables each node to selectively retrieve relevant geometric information across multiple spatial scales.

\paragraph{Feature adapter.}
In addition to the geometric adaptation residual, we incorporate a feature-level adapter to refine node representations:
\begin{equation}
\Delta \mathbf{h}_{i,\mathrm{feat}}^{(l)} =
\mathbf{W}_2^{(l)}
\sigma(
\mathbf{W}_1^{(l)}
\mathbf{h}_i^{(l-1)}
).
\end{equation}
The final node update combines the pre-trained representation with the
feature-level residual and the geometric adaptation residual:
\begin{equation}
\tilde{\mathbf{h}}_i^{(l)} =
\mathbf{h}_i^{(l)}
+
\gamma_0^{(l)}
\Delta \mathbf{h}_{i,\mathrm{feat}}^{(l)}
+
\gamma_1^{(l)}
\Delta \mathbf{h}_{i,\mathrm{geo}}^{(l)}.
\end{equation}
This gated formulation combines the pre-trained representation,
feature-level adaptation, and multi-scale conformer geometry while
allowing their contributions to be learned separately.

\section{Experiments}

\subsection{Dataset}

\paragraph{Dataset collection.}
In this work, we focus on the blood--brain barrier permeability prediction task and follow the dataset construction protocol in~\cite{Vinh2025}. 
We adopt the BBBP dataset collected from the B3DB database~\cite{Meng2021}, which aggregates molecular records from multiple peer-reviewed publications and public sources. 
The original dataset consists of 7,807 compounds, including 4,956 BBB-permeable (BBB+) and 2,851 non-permeable (BBB-) molecules. 
After preprocessing and quality control, the final curated dataset contains 3,832 samples {(2,446 BBB+, 1,386 BBB$-$, ratio 1.76:1)}. {The architecture is task-independent, but the empirical evidence in this paper is limited to BBBP.}

\paragraph{Data curation.}
To ensure data quality and reliability, we apply a standardized chemical data curation pipeline following~\cite{Vinh2025}. 
The pipeline consists of four stages: (i) validation, (ii) cleaning, (iii) normalization, and (iv) final verification. 
First, all molecular structures are standardized into canonical SMILES format. 
During validation, invalid samples, such as inorganic compounds, mixtures, and organometallics, are removed. 
In the cleaning stage, molecules containing metal ions are discarded, and charged molecules are neutralized when appropriate. 
Next, normalization reduces structural redundancy by transforming stereoisomers and tautomers into canonical forms, mitigating duplication caused by equivalent representations. 
Finally, a verification step removes duplicate entries based on molecular identifiers and resolves label inconsistencies. 
Additional cross-referencing with PubChem and ChEMBL further ensures structural correctness and label reliability.

\paragraph{Dataset splitting. }We evaluate model performance under both random and scaffold splitting strategies. For scaffold splitting, molecules are partitioned according to their Bemis--Murcko scaffolds~\cite{bemis1996properties}, ensuring mutually exclusive scaffold groups across training, validation, and test sets, thereby providing a more realistic evaluation of generalization to unseen chemical structures. In both settings, 10\% of the dataset is reserved for testing, while the remaining samples are split into training and validation sets at a 90:10 ratio. Across five runs, both splitting strategies broadly maintain similar BBB+/BBB- proportions, although scaffold splitting introduces larger variation in class distribution and performance.

\subsection{Baselines}

We compare the proposed method against both full fine-tuning and a diverse set of parameter-efficient fine-tuning approaches. 
Specifically, we consider representative PEFT methods including Adapter~\cite{houlsby2019parameter}, LoRA~\cite{hu2022lora}, GPF and GPF-plus~\cite{fang2023universal}, and AdapterGNN~\cite{Li2024}. 
Following~\cite{Li2024}, to evaluate robustness across different pre-training objectives, we further incorporate three widely used self-supervised pre-training strategies, all based on a pre-trained Graph Isomorphism Network (GIN)~\cite{xu2018how} backbone: SimGRACE~\cite{xia2022simgrace}, EdgePred~\cite{Hu*2020Strategies}, and ContextPred~\cite{Hu*2020Strategies}. We note that the experimental focus is on parameter-efficient adaptation under matched pre-trained backbones.


\subsection{Implementation Details}

All models are trained using the AdamW optimizer with an initial learning rate of $3\times10^{-4}$ and a batch size of 64 for 150 epochs. To ensure a fair comparison, we apply consistent hyperparameter settings across all parameter-efficient fine-tuning methods. We employ the GIN as the backbone for auxiliary geometric graph encoders. Following~\cite{Li2024}, we freeze the pre-trained backbone except for its
batch-normalization parameters, while training BBBP-GeoPEFT and the
task-specific prediction head. Unless otherwise specified, the rank of the low-rank factorization is fixed at $r=16$. For multi-scale graph construction, {we use distance cutoffs $\{2, 2.5, 3, 3.5, 4\}\,\text{\AA}$, a set also used by Mol-TDL~\cite{shen2024molecular} to model covalent and non-covalent interactions across scales.} For each canonical SMILES, explicit hydrogens were added and 10 conformers were generated using RDKit ETKDGv3~\cite{landrum2013rdkit} with random seed 42. Conformers were optimized using MMFF94 (or UFF when unavailable), and the lowest-energy conformer was selected before removing hydrogens and constructing the distance and line graphs.

We evaluate model performance using standard classification metrics, including the Area Under the Receiver Operating Characteristic Curve (ROC-AUC), Precision-Recall Area Under the Curve (PR-AUC), Accuracy, and F1-score. To assess robustness, each experiment is repeated over 5 seeds, and we report the mean and standard deviation. All experiments are conducted on NVIDIA RTX A5000 GPUs with 24GB of memory.


\subsection{Model Evaluation}

\begin{table*}[!ht]
\centering
\caption{Comparison with full fine-tuning and representative PEFT methods on the BBBP dataset under random splitting strategy. Results are reported as mean $\pm$ standard deviation over 5 runs, in percentage. Best and second-best results within each pre-training group are highlighted in \textcolor{red}{red} and \textcolor{blue}{blue}, respectively. Lower Avg.\ Rank is better.}
\label{tab:baselines_peft_random_split}
\setlength{\tabcolsep}{8pt}
\begin{tabular}{@{}llcccc|c@{}}
\toprule
\textbf{Pre-training}
  & \textbf{Method}
  & \textbf{ROC-AUC $\uparrow$}
  & \textbf{PR-AUC $\uparrow$}
  & \textbf{Accuracy $\uparrow$}
  & \textbf{F1 $\uparrow$}
  & \textbf{Avg.\ Rank $\downarrow$} \\
\midrule

\multirow{7}{*}{SimGRACE}
  & Full FT                          & \val{85.4}{0.2}       & \val{91.4}{0.3}       & \val{77.0}{0.6}       & \val{82.8}{0.4}       & 4.38 \\
  & Adapter                          & \secondval{85.6}{0.5} & \secondval{91.7}{0.5} & \val{76.3}{0.7}       & \val{82.1}{0.8}       & 4.63 \\
  & LoRA                             & \val{84.9}{0.8}       & \val{91.1}{0.7}       & \val{76.9}{1.7}       & \val{82.6}{1.4}       & 5.50 \\
  & GPF                              & \val{82.5}{0.6}       & \val{87.7}{0.6}       & \val{77.3}{0.8}       & \secondval{83.3}{0.5} & 4.75 \\
  & GPF-plus                         & \val{84.0}{1.0}       & \val{88.8}{1.1}       & \val{77.2}{1.1}       & \val{83.2}{0.9}       & 4.75 \\
  & AdapterGNN                       & \secondval{85.6}{1.2} & \val{91.4}{1.0}       & \secondval{77.4}{1.1} & \val{82.9}{0.8}       & 3.00 \\
  & \textbf{BBBP-GeoPEFT}  & \topval{87.1}{1.1}    & \topval{91.8}{0.8}    & \topval{79.6}{1.0}    & \topval{83.7}{0.8}    & \bestrank{1.00} \\
\midrule

\multirow{7}{*}{EdgePred}
  & Full FT                          & \secondval{87.3}{0.5} & \topval{92.8}{0.3}    & \val{77.1}{0.7}       & \val{83.3}{0.4}       & 2.75 \\
  & Adapter                          & \val{85.6}{0.6}       & \val{91.9}{0.3}       & \val{76.2}{0.6}       & \val{82.3}{0.8}       & 5.00 \\
  & LoRA                             & \val{86.8}{0.2}       & \secondval{92.4}{0.2} & \val{77.5}{0.4}       & \secondval{83.4}{0.3} & 3.00 \\
  & GPF                              & \val{72.0}{1.1}       & \val{81.4}{1.1}       & \val{66.6}{1.6}       & \val{77.1}{1.5}       & 6.75 \\
  & GPF-plus                         & \val{72.3}{4.1}       & \val{80.4}{2.9}       & \val{69.1}{3.3}       & \val{80.1}{1.5}       & 6.25 \\
  & AdapterGNN                       & \topval{87.4}{0.6}    & \topval{92.8}{0.4}    & \secondval{77.8}{1.1} & \topval{83.9}{0.5}    & \bestrank{1.38} \\
  & \textbf{BBBP-GeoPEFT}  & \secondval{87.3}{0.3} & \val{92.1}{0.4}       & \topval{78.8}{1.0}    & \val{83.2}{0.7}       & 2.88 \\
\midrule

\multirow{7}{*}{ContextPred}
  & Full FT                          & \secondval{86.8}{0.7} & \secondval{92.6}{0.4} & \secondval{77.6}{1.0} & \secondval{83.1}{0.6} & 2.25 \\
  & Adapter                          & \val{86.2}{0.7}       & \val{92.3}{0.4}       & \val{76.9}{0.9}       & \val{83.0}{0.7}       & 4.25 \\
  & LoRA                             & \val{86.6}{0.5}       & \val{92.4}{0.4}       & \val{76.7}{0.9}       & \secondval{83.1}{0.5} & 3.38 \\
  & GPF                              & \val{78.8}{0.6}       & \val{86.3}{0.6}       & \val{70.7}{1.8}       & \val{80.7}{0.6}       & 7.00 \\
  & GPF-plus                         & \val{79.1}{1.0}       & \val{86.7}{0.8}       & \val{72.0}{2.4}       & \val{81.4}{1.2}       & 6.00 \\
  & AdapterGNN                       & \val{86.5}{0.4}       & \secondval{92.6}{0.2} & \val{76.1}{0.6}       & \val{82.6}{0.3}       & 4.13 \\
  & \textbf{BBBP-GeoPEFT}  & \topval{88.1}{0.7}    & \topval{92.8}{0.4}    & \topval{79.9}{0.9}    & \topval{84.0}{1.0}    & \bestrank{1.00} \\

\bottomrule
\end{tabular}
\end{table*}

\subsubsection{Comparison with Full Fine-Tuning and PEFT Baselines under Random Split.}

Table~\ref{tab:baselines_peft_random_split} compares BBBP-GeoPEFT with representative PEFT baselines under the random split setting. Overall, BBBP-GeoPEFT achieves the best or second-best performance across most metrics and pre-training objectives. Under SimGRACE and ContextPred pre-training, BBBP-GeoPEFT consistently achieves the strongest overall performance, reaching up to 88.1\% ROC-AUC and 84.0\% F1-score, while updating only a small fraction of model parameters compared with full fine-tuning.
Under the EdgePred backbone, BBBP-GeoPEFT remains competitive and achieves the highest Accuracy, although AdapterGNN obtains the best average ranking and stronger PR-AUC and F1-score. This may be because EdgePred pre-training already captures informative local bond and edge-level structural patterns, reducing the additional benefit of geometry-informed adaptation.
Across all settings, adapter-based methods often outperform prompt-based approaches such as GPF and GPF-plus. Moreover, incorporating multi-scale geometric information generally leads to stronger performance than generic PEFT baselines in these settings.

\subsubsection{Comparison with Full Fine-Tuning and PEFT Baselines under Scaffold Split.}

Table~\ref{tab:baselines_peft_scaffold_split} reports the results under the more challenging scaffold split setting. Compared with random split, all methods show lower performance and larger standard deviations, reflecting the difficulty of generalizing to unseen molecular scaffolds. Nevertheless, BBBP-GeoPEFT consistently maintains competitive performance across different pre-training objectives.
Under SimGRACE pre-training, BBBP-GeoPEFT achieves the best average ranking and improves over most baselines. Under ContextPred pre-training, BBBP-GeoPEFT achieves the strongest overall results, reaching 86.2\% ROC-AUC, 91.3\% PR-AUC, 78.9\% Accuracy, and 84.0\% F1-score. In contrast, under the EdgePred backbone, BBBP-GeoPEFT remains competitive but does not consistently outperform Full FT and AdapterGNN across all metrics, suggesting that EdgePred already encodes informative local structural patterns.
Similar to the random split setting, prompt-based methods perform substantially worse than adapter-based approaches. {Overall, BBBP-GeoPEFT achieves the best average ranking under SimGRACE and ContextPred. Under EdgePred, it remains competitive with full fine-tuning, although it does not consistently outperform AdapterGNN.}

\begin{table*}[!ht]
\centering
\caption{Comparison with full fine-tuning and representative PEFT methods on the BBBP dataset under scaffold-based splitting strategy. Results are reported as mean $\pm$ standard deviation over 5 runs, in percentage. Best and second-best results within each pre-training group are highlighted in \textcolor{red}{red} and \textcolor{blue}{blue}, respectively. Lower Avg.\ Rank is better.}
\label{tab:baselines_peft_scaffold_split}
\setlength{\tabcolsep}{8pt}
\begin{tabular}{@{}llcccc|c@{}}
\toprule
\textbf{Pre-training}
  & \textbf{Method}
  & \textbf{ROC-AUC $\uparrow$}
  & \textbf{PR-AUC $\uparrow$}
  & \textbf{Accuracy $\uparrow$}
  & \textbf{F1 $\uparrow$}
  & \textbf{Avg.\ Rank $\downarrow$} \\
\midrule

\multirow{7}{*}{SimGRACE}
  & Full FT                          & \val{82.9}{4.8}       & \val{89.0}{4.5}       & \val{75.6}{6.2}       & \val{82.2}{5.1}       & 4.75 \\
  & Adapter                          & \secondval{84.4}{4.9} & \val{89.7}{4.4}       & \secondval{77.2}{4.9} & \topval{82.8}{4.7}    & 2.25 \\
  & LoRA                             & \val{83.1}{6.1}       & \val{89.3}{5.6}       & \val{76.0}{5.4}       & \val{81.8}{5.1}       & 4.50 \\
  & GPF                              & \val{79.8}{6.3}       & \val{86.3}{6.4}       & \val{75.0}{6.6}       & \val{82.0}{5.6}       & 6.25 \\
  & GPF-plus                         & \val{80.8}{6.5}       & \val{86.8}{6.4}       & \val{74.2}{6.8}       & \val{81.4}{5.8}       & 6.50 \\
  & AdapterGNN                       & \secondval{84.4}{4.9} & \secondval{90.0}{5.1} & \val{77.0}{5.1}       & \secondval{82.5}{5.2} & 2.63 \\
  & \textbf{BBBP-GeoPEFT}  & \topval{85.0}{5.2}    & \topval{90.2}{5.1}    & \topval{78.2}{4.9}    & \topval{82.8}{5.0}    & \bestrank{1.13} \\
\midrule

\multirow{7}{*}{EdgePred}
  & Full FT                          & \secondval{87.7}{2.3} & \secondval{92.5}{2.9} & \topval{79.7}{4.1}    & \topval{85.4}{3.7}    & \bestrank{1.63} \\
  & Adapter                          & \val{86.3}{3.1}       & \val{91.5}{3.7}       & \val{78.6}{3.4}       & \val{84.1}{3.9}       & 5.00 \\
  & LoRA                             & \val{87.4}{2.3}       & \val{92.0}{3.3}       & \secondval{79.2}{3.4} & \secondval{84.9}{3.3} & 3.13 \\
  & GPF                              & \val{73.1}{6.6}       & \val{82.0}{7.1}       & \val{68.1}{6.0}       & \val{76.4}{6.8}       & 7.00 \\
  & GPF-plus                         & \val{74.8}{4.8}       & \val{82.7}{5.3}       & \val{72.1}{5.1}       & \val{80.7}{4.1}       & 6.00 \\
  & AdapterGNN                       & \topval{87.8}{2.7}    & \topval{92.6}{3.2}    & \val{78.7}{3.5}       & \val{84.3}{3.9}       & 2.50 \\
  & \textbf{BBBP-GeoPEFT}  & \secondval{87.7}{2.3} & \val{92.1}{3.2}       & \secondval{79.2}{4.8} & \val{84.5}{4.2}       & 2.75 \\
\midrule

\multirow{7}{*}{ContextPred}
  & Full FT                          & \val{84.6}{5.6}       & \val{90.6}{5.2}       & \val{77.6}{5.8}       & \val{83.4}{5.0}       & 3.75 \\
  & Adapter                          & \val{84.6}{4.2}       & \val{90.4}{4.6}       & \val{77.3}{4.9}       & \val{83.4}{4.5}       & 4.50 \\
  & LoRA                             & \val{85.0}{3.9}       & \val{90.9}{3.9}       & \secondval{78.1}{4.2} & \secondval{83.9}{4.0} & 2.50 \\
  & GPF                              & \val{71.4}{9.2}       & \val{81.3}{8.3}       & \val{69.7}{5.9}       & \val{80.5}{4.3}       & 7.00 \\
  & GPF-plus                         & \val{72.8}{6.4}       & \val{81.9}{7.3}       & \val{70.7}{6.7}       & \val{80.7}{4.8}       & 6.00 \\
  & AdapterGNN                       & \secondval{85.3}{4.9} & \secondval{91.2}{4.5} & \val{77.4}{5.4}       & \val{83.2}{5.3}       & 3.25 \\
  & \textbf{BBBP-GeoPEFT}  & \topval{86.2}{4.6}    & \topval{91.3}{4.3}    & \topval{78.9}{5.3}    & \topval{84.0}{4.9}    & \bestrank{1.00} \\

\bottomrule
\end{tabular}
\end{table*}

\subsection{Model Analysis}

\subsubsection{Parameter Efficiency Analysis}

\begin{table}[!ht]
\centering
\caption{Trainable parameter comparison on BBBP. The percentage is computed relative to full fine-tuning.}
\label{tab:trainable_params}
\begin{tabular}{@{}lrr@{}}
\toprule
\textbf{Method} & \textbf{Trainable Params} & \textbf{\% of Full FT} \\
\midrule
Full FT     & 1,858,201 & 100.00\% \\
Adapter     & 114,471   & 6.2\% \\
LoRA        & 90,301    & 4.9\% \\
GPF         & 601       & 0.03\% \\
GPF-plus    & 3,306     & 0.18\% \\
AdapterGNN  & 106,961    & 5.8\% \\
\midrule
BBBP-GeoPEFT & 188,515   & 10.1\% \\
\bottomrule
\end{tabular}
\end{table}

Table~\ref{tab:trainable_params} compares the number of trainable parameters across different PEFT methods. As expected, PEFT approaches substantially reduce the parameter cost compared to full fine-tuning. Prompt-based methods, such as GPF and GPF-plus, achieve the smallest parameter footprint, but exhibit noticeable performance degradation in Tables~\ref{tab:baselines_peft_random_split} and~\ref{tab:baselines_peft_scaffold_split}. In contrast, adapter-based methods provide a better balance between efficiency and predictive performance.
BBBP-GeoPEFT introduces a moderate increase in trainable parameters (10.1\%) due to the auxiliary geometric graph encoders and
node-wise cutoff attention module. Despite this overhead, it consistently achieves superior or competitive performance across multiple pre-trained backbones and splitting strategies. These results suggest that incorporating conformer-derived graph representations can significantly improve representation quality while maintaining the efficiency advantages of parameter-efficient adaptation.

\subsubsection{Effect of Low-Rank Factorization Dimension}

\begin{figure}[!ht]
    \centering
    \Description{Ranking impact. }\includegraphics[width=0.475\textwidth]{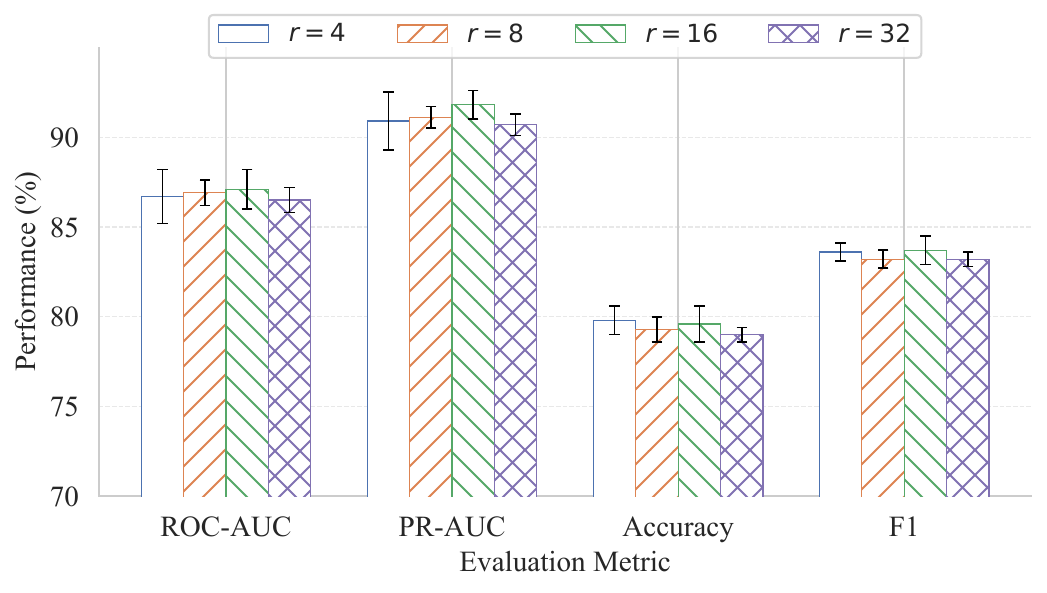}
    \caption{Effect of the low-rank factorization dimension $r$ on BBBP prediction using the SimGRACE-pre-trained backbone. Results are reported as mean $\pm$ standard deviation over 5 runs. BBBP-GeoPEFT achieves stable performance across different ranks, with the best results observed at $r=16$.
    }
\label{fig:rank_comparison}
\end{figure}

We analyze the impact of the low-rank factorization dimension $r$ on model performance, as shown in Figure~\ref{fig:rank_comparison}. We evaluate BBBP-GeoPEFT with $r \in \{4, 8, 16, 32\}$ using the SimGRACE-pre-trained backbone. Across all evaluation metrics, including ROC-AUC, PR-AUC, Accuracy, and F1-score, the model demonstrates stable performance with only minor variation across different ranks.
Specifically, increasing the rank from $r=4$ to $r=16$ leads to consistent but marginal improvements, with the best performance achieved at $r=16$ (87.1\% ROC-AUC, 91.8\% PR-AUC, and 83.7\% F1). Further increasing the rank to $r=32$ does not yield additional gains and slightly degrades performance on some metrics. This suggests that a moderate rank is sufficient to capture the necessary adaptation capacity, while larger ranks may introduce redundancy or overfitting.


\subsubsection{Statistical Significance Analysis}

\begin{table}[t]
\centering
\caption{Paired significance analysis of BBBP-GeoPEFT against baselines across 5 runs under random splitting strategy. $\Delta$ denotes the mean difference between BBBP-GeoPEFT and the baseline. $p_t$ denotes the two-sided paired $t$-test $p$-value.}
\label{tab:paired_tests_random_split}
\scriptsize
\setlength{\tabcolsep}{3.5pt}
\renewcommand{\arraystretch}{1.08}
\resizebox{0.475\textwidth}{!}{
\begin{tabular}{@{}llcc cc cc cc@{}}
\toprule
\multirow{2}{*}{\textbf{Pre-training}} 
& \multirow{2}{*}{\textbf{Baseline}}
& \multicolumn{2}{c}{\textbf{ROC-AUC}}
& \multicolumn{2}{c}{\textbf{PR-AUC}}
& \multicolumn{2}{c}{\textbf{Accuracy}}
& \multicolumn{2}{c}{\textbf{F1}} \\
\cmidrule(lr){3-4}
\cmidrule(lr){5-6}
\cmidrule(lr){7-8}
\cmidrule(lr){9-10}
& & $\Delta$ & $p_t$ & $\Delta$ & $p_t$ & $\Delta$ & $p_t$ & $\Delta$ & $p_t$ \\
\midrule

\multirow{6}{*}{SimGRACE}
& Full FT    & +1.78 & 0.0296 & +0.35 & 0.4498 & +2.66 & 0.0145 & +0.89 & 0.1631 \\
& Adapter    & +1.55 & 0.0056 & +0.12 & 0.7314 & +3.34 & 0.0011 & +1.57 & 0.0326 \\
& LoRA       & +2.20 & 0.0010 & +0.70 & 0.0397 & +2.76 & 0.0287 & +1.13 & 0.1969 \\
& AdapterGNN & +1.54 & 0.2010 & +0.43 & 0.5986 & +2.19 & 0.0724 & +0.82 & 0.3247 \\
& GPF        & +4.69 & 0.0007 & +4.13 & 0.0011 & +2.30 & 0.0206 & +0.37 & 0.4840 \\
& GPF-plus   & +3.17 & 0.0006 & +3.00 & 0.0003 & +2.40 & 0.0194 & +0.44 & 0.5110 \\
\midrule

\multirow{6}{*}{EdgePred}
& Full FT    & -0.04 & 0.8902 & -0.76 & 0.0033 & +1.67 & 0.0896 & -0.15 & 0.7874 \\
& Adapter    & +1.67 & 0.0108 & +0.20 & 0.4881 & +2.61 & 0.0016 & +0.87 & 0.0136 \\
& LoRA       & +0.50 & 0.0250 & -0.33 & 0.2492 & +1.31 & 0.0089 & -0.23 & 0.4585 \\
& AdapterGNN & -0.11 & 0.5162 & -0.75 & 0.0209 & +1.04 & 0.2177 & -0.69 & 0.0970 \\
& GPF        & +15.30 & < 0.0001& +10.71 & < 0.0001& +12.19 & 0.0005 & +6.10 & 0.0009 \\
& GPF-plus   & +14.97 & 0.0012 & +11.73 & 0.0008 & +9.74 & 0.0037 & +3.07 & 0.0254 \\
\midrule

\multirow{6}{*}{ContextPred}
& Full FT    & +1.35 & 0.0682 & +0.17 & 0.5973 & +2.30 & 0.0313 & +0.88 & 0.2665 \\
& Adapter    & +1.95 & 0.0214 & +0.52 & 0.1038 & +3.03 & 0.0076 & +0.96 & 0.1416 \\
& LoRA       & +1.49 & 0.0384 & +0.38 & 0.2883 & +3.18 & 0.0063 & +0.93 & 0.0999 \\
& AdapterGNN & +1.60 & 0.0132 & +0.20 & 0.4246 & +3.75 & 0.0033 & +1.40 & 0.0466 \\
& GPF        & +9.26 & < 0.0001& +6.46 & < 0.0001& +9.17 & 0.0007 & +3.27 & 0.0037 \\
& GPF-plus   & +8.97 & < 0.0001& +6.13 & < 0.0001& +7.87 & 0.0035 & +2.65 & 0.0313 \\

\bottomrule
\end{tabular}
}
\end{table}

\paragraph{Random split. }We evaluate the robustness of BBBP-GeoPEFT using paired two-sided $t$-tests across five runs under the random split setting. As shown in Table~\ref{tab:paired_tests_random_split}, BBBP-GeoPEFT consistently achieves positive improvements across most pre-training objectives and evaluation metrics. Under SimGRACE pre-training, BBBP-GeoPEFT shows statistically significant gains over several PEFT baselines, particularly on ROC-AUC and Accuracy. For example, compared with LoRA, BBBP-GeoPEFT improves ROC-AUC by $+2.20$ ($p=0.0010$) and Accuracy by $+2.76$ ($p=0.0287$). Under ContextPred pre-training, BBBP-GeoPEFT consistently outperforms AdapterGNN, achieving significant improvements on ROC-AUC ($+1.60$, $p=0.0132$), Accuracy ($+3.75$, $p=0.0033$), and F1 ($+1.40$, $p=0.0466$). Although the gains against strong baselines such as Full FT and AdapterGNN under EdgePred are generally smaller and not always statistically significant, BBBP-GeoPEFT remains competitive while substantially outperforming prompt-based methods such as GPF and GPF-plus across all metrics and pre-training objectives. 
Overall, these comparisons provide evidence that attention-based
geometric adaptation can improve parameter-efficient fine-tuning,
although the gains depend on the pre-training objective and metric.

\begin{table}[t]
\centering
\caption{Paired significance analysis of BBBP-GeoPEFT against baselines across 5 runs under scaffold splitting strategy. $\Delta$ denotes the mean difference between BBBP-GeoPEFT and the baseline. $p_t$ denotes the two-sided paired $t$-test $p$-value.}
\label{tab:paired_tests_scaffold_split}
\scriptsize
\setlength{\tabcolsep}{3.5pt}
\renewcommand{\arraystretch}{1.08}
\resizebox{0.475\textwidth}{!}{
\begin{tabular}{@{}llcc cc cc cc@{}}
\toprule
\multirow{2}{*}{\textbf{Pre-training}}
& \multirow{2}{*}{\textbf{Baseline}}
& \multicolumn{2}{c}{\textbf{ROC-AUC}}
& \multicolumn{2}{c}{\textbf{PR-AUC}}
& \multicolumn{2}{c}{\textbf{Accuracy}}
& \multicolumn{2}{c}{\textbf{F1}} \\
\cmidrule(lr){3-4}
\cmidrule(lr){5-6}
\cmidrule(lr){7-8}
\cmidrule(lr){9-10}
& & $\Delta$ & $p_t$ & $\Delta$ & $p_t$ & $\Delta$ & $p_t$ & $\Delta$ & $p_t$ \\
\midrule

\multirow{6}{*}{SimGRACE}
& Full FT    & +2.05 & 0.2226 & +1.14 & 0.4010 & +2.61 & 0.1558 & +0.61 & 0.5899 \\
& Adapter    & +0.61 & 0.1744 & +0.48 & 0.3776 & +1.04 & 0.1477 & +0.01 & 0.9752 \\
& LoRA       & +1.86 & 0.0315 & +0.86 & 0.2755 & +2.25 & 0.0054 & +1.06 & 0.2480 \\
& AdapterGNN & +0.57 & 0.2044 & +0.19 & 0.3274 & +1.20 & 0.0866 & +0.31 & 0.6638 \\
& GPF        & +5.15 & 0.0017 & +3.87 & 0.0051 & +3.24 & 0.0859 & +0.79 & 0.5598 \\
& GPF-plus   & +4.21 & 0.0104 & +3.34 & 0.0623 & +4.02 & 0.0343 & +1.47 & 0.2369 \\
\midrule

\multirow{6}{*}{EdgePred}
& Full FT    & +0.01 & 0.9926 & -0.40 & 0.6743 & -0.52 & 0.7815 & -0.90 & 0.5572 \\
& Adapter    & +1.35 & 0.0555 & +0.59 & 0.3693 & +0.63 & 0.5796 & +0.48 & 0.6901 \\
& LoRA       & +0.28 & 0.4398 & +0.04 & 0.9607 & +0.05 & 0.9650 & -0.39 & 0.7450 \\
& AdapterGNN & -0.08 & 0.9061 & -0.59 & 0.4518 & +0.53 & 0.6418 & +0.21 & 0.8494 \\
& GPF        & +14.64 & 0.0027 & +10.02 & 0.0050 & +11.12 & 0.0047 & +8.14 & 0.0164 \\
& GPF-plus   & +12.92 & 0.0005 & +9.34 & 0.0008 & +7.16 & 0.0055 & +3.83 & 0.0452 \\
\midrule

\multirow{6}{*}{ContextPred}
& Full FT    & +1.60 & 0.0527 & +0.68 & 0.2089 & +1.30 & 0.0778 & +0.56 & 0.3784 \\
& Adapter    & +1.58 & 0.0953 & +0.93 & 0.1532 & +1.56 & 0.0582 & +0.58 & 0.3991 \\
& LoRA       & +1.15 & 0.1443 & +0.39 & 0.5014 & +0.83 & 0.2232 & +0.13 & 0.7882 \\
& AdapterGNN & +0.92 & 0.0043 & +0.12 & 0.3596 & +1.46 & 0.0055 & +0.81 & 0.0436 \\
& GPF        & +14.76 & 0.0317 & +10.04 & 0.0393 & +9.19 & 0.0035 & +3.51 & 0.0377 \\
& GPF-plus   & +13.44 & 0.0144 & +9.36 & 0.0268 & +8.25 & 0.0122 & +3.24 & 0.0482 \\

\bottomrule
\end{tabular}}
\end{table}

\paragraph{Scaffold split. }We further evaluate the robustness of BBBP-GeoPEFT using paired two-sided $t$-tests across five runs under the scaffold splitting strategy. As shown in Table~\ref{tab:paired_tests_scaffold_split}, BBBP-GeoPEFT consistently achieves positive improvements across most pre-training objectives and evaluation metrics. Under SimGRACE pre-training, BBBP-GeoPEFT shows statistically significant gains over LoRA on ROC-AUC ($+1.86$, $p=0.0315$) and Accuracy ($+2.25$, $p=0.0054$), while also substantially outperforming prompt-based methods such as GPF and GPF-plus. Under ContextPred pre-training, BBBP-GeoPEFT consistently outperforms AdapterGNN, with significant improvements on ROC-AUC ($+0.92$, $p=0.0043$), Accuracy ($+1.46$, $p=0.0055$), and F1 ($+0.81$, $p=0.0436$). In contrast, the improvements over strong baselines such as Full FT and AdapterGNN under EdgePred are generally smaller and not always statistically significant, indicating that the effectiveness of geometry-informed adaptation depends on the pre-training objective. Nevertheless, BBBP-GeoPEFT remains competitive with strong PEFT and full fine-tuning baselines while consistently achieving large gains over prompt-based methods across all settings.

\section{Conclusion}

In this work, we propose \textbf{BBBP-GeoPEFT}, a geometry-informed parameter-efficient fine-tuning framework for blood--brain barrier permeability prediction. 
BBBP-GeoPEFT augments pre-trained molecular GNNs with multi-scale
conformer geometry represented by distance-based graphs and their
corresponding line graphs. The two graph families model spatial atom
connectivity and second-order edge interactions, respectively. Shared
auxiliary geometric graph encoders produce cutoff-specific
representations, and node-wise cutoff attention aggregates them into a
geometric adaptation residual at each backbone layer. This formulation
incorporates conformer-derived information while updating only a limited
subset of model parameters.
{Across both split settings, BBBP-GeoPEFT achieves the best average ranking under SimGRACE and ContextPred while updating only 10.1\% of the model parameters. Under EdgePred, it remains competitive with full fine-tuning, although it does not consistently outperform AdapterGNN.} In addition, the rank-sensitivity analysis shows that performance
remains stable across the evaluated low-rank factorization dimensions. Overall, this work provides a practical approach for incorporating {these geometric signals} into pre-trained molecular models.

Despite these promising results, several limitations remain. The proposed framework relies on RDKit-generated conformers, which may not fully capture true molecular geometry, particularly for flexible molecules. In addition, while parameter-efficient, the method introduces additional computational cost due to multi-scale
graph construction and auxiliary graph encoding.
Future work will focus on improving geometric representations and extending the modeling capacity. In particular, incorporating more accurate conformer generation and exploring richer structural features, such as angular or higher-order interactions, may further enhance performance. We also plan to investigate more advanced parameter-efficient adaptation strategies to improve efficiency and generalization.

\section{Acknowledgments}

This study was supported in part by the Endeavour Fund -- Smart Ideas initiative from the New Zealand Ministry of Business, Innovation and Employment (MBIE) under contract RTVU2301, the Royal Society of New Zealand through the Catalyst: Seeding Grant (CSG-VUW2503), and the Explorer Grant 25/910/A from the Health Research Council of New Zealand.

\printbibliography

\end{document}